\documentclass{article}

\usepackage[preprint]{neurips_2023}

\usepackage[utf8]{inputenc} % allow utf-8 input
\usepackage[T1]{fontenc}    % use 8-bit T1 fonts
\usepackage{hyperref}       % hyperlinks
\usepackage{url}            % simple URL typesetting
\usepackage{booktabs}       % professional-quality tables
\usepackage{amsfonts}       % blackboard math symbols
\usepackage{nicefrac}       % compact symbols for 1/2, etc.
\usepackage{microtype}      % microtypography
\usepackage{xcolor}         % colors
\usepackage[noline,noend, boxruled]{algorithm2e}
\usepackage{amsmath}
\usepackage{amssymb}
\usepackage{caption}
\usepackage{subcaption}
\usepackage{graphicx}

\newcommand{\LexV}{\textsc{Lex}_{\textsc{V}}}
\newcommand{\LexN}{\textsc{Lex}_{\textsc{N}}}
\newcommand{\Phon}{\textsc{Phon}}
\newcommand{\Context}{\textsc{Context}}
\newcommand{\Visual}{\textsc{Visual}}
\newcommand{\Motor}{\textsc{Motor}}

\newcommand{\Lang}{\textsc{Lang}}

\title{The Architecture of a Biologically Plausible \\
Language Organ}

\author{%
  Daniel Mitropolsky \\
  Department of Computer Science\\
  Columbia University\\
  New York, NY \\
  \texttt{mitropolsky@cs.columbia.edu} \\
  \And
  Christos H. Papadimitriou \\
  Department of Computer Science\\
  Columbia University\\
  New York, NY \\
  \texttt{christos@columbia.edu} \\
}

\begin{document}

\maketitle

\begin{abstract}
 We present a simulated biologically plausible \emph{language organ}, made up of stylized but realistic neurons, synapses, brain areas, plasticity, and a simplified model of sensory perception.  We show through experiments that this model succeeds in an important early step in language acquisition: the learning of nouns, verbs, and their meanings, from the grounded input of only a modest number of sentences. Learning in this system is achieved through Hebbian plasticity, and \emph{without} backpropagation. Our model goes beyond a {\em parser} previously designed in a similar environment, with the critical addition of a biologically plausible account for how language can be acquired in the infant's brain, not just processed by a mature brain. %Our work is a contribution to the study of the neural basis of cognition and language in particular, but also may be of interest to artificial intelligence where more human-like learning processes may be key to overcoming current weaknesses of even the most powerful AI artifacts.
\end{abstract}

\section{Introduction}
It is beyond doubt that cognitive phenomena such as language, reasoning, and planning are the direct product of the activity of neurons and synapses.  However, there is no extant overarching theory explaining exactly how this is done.  %Bridging the  the gap between the neuron and the brain. This is a gap not just of scale, but also of scientific tradition and experimental methodology. 
In the words of 2004 Nobel laureate Richard Axel \citep{AxelNeuron2018} \emph{``We do not have a Logic for the transformation of neural activity into thought and action.''}   Making progress on this open question, often called the \emph{bridging problem} \citep{PapadimitriouFriederici}, is identified by Axel (ibid.) as the most important challenge facing neuroscience today.  

In recent years, a computational approach to the bridging problem has been undertaken. In \citep{PNAS}, a computational system called the Assembly Calculus was proposed, based on a simplified mathematical model of spiking neurons and synapses, which reflects the basic elements and tenets of Neuroscience: brain areas, excitatory neurons, local inhibition, plasticity (see the next section for a detailed description of the enhanced version of this model used here). Within this framework, neuromorphic computational systems simulating certain large-scale cognitive phenomena were implemented: a system for planning in the blocks world \citep{Planning}; a system for learning to classify representations through few-shot training \citep{Dabagia}; and, perhaps more surprisingly, a system for \emph{parsing sentences} in English and other languages \citep{Parser,CenterEmbedding}. 

We believe that pursuing this research program of constructing more and more ambitious neuromorphic artifacts simulating cognitive phenomena is important, for at least two reasons.  First, each step on this path entails concrete progress in the bridging problem, as more and more advanced domains of cognition are explored through artifacts consisting of reasonably realistic and brain-like, if stylized, systems of neurons and synapses.  Second, further progress in this direction may be of interest to Artificial Intelligence: Despite amazing advances over the past ten years, arguably AI still lags behind human brains in several important dimensions: grounded language, continual learning, originality and inventiveness, emotional and social intelligence, and energy usage.  Creating intelligent artifacts that are more brain-like, and rely on modes of learning other than backpropagation, may eventually point to new possible avenues of progress for AI.

The biologically plausible parser of \citet{Parser} includes a lexicon containing neural representations of words. It is assumed that each neural representation of a word is wired so that, when excited by an outside stimulus, it sets in motion specific neural activities inhibiting and/or disinhibiting remote brain areas that are associated with the word's syntactic role (verb, subject, etc.). This works fine for the purposes of the parser, except that it leaves open perhaps the most important questions: How are these word representations created?  How are these neural activities set up in the infant brain and how are they associated with the representation of each word, thus implementing the word's part of speech?  And how are those other brain areas labeled with the appropriate syntactic roles?  In other words, \emph{how is language acquired in the human brain?} 

This is the question we set out to answer in this paper.

We seek to create a \emph{neuromorphic language organ:} a tabula rasa of neural components --- roughly, a collection of brain areas with randomly connected neurons, with certain additional neural populations, all consistent with basic Neuroscience and plausibly set in place during the infant's development --- which, upon the input of modest amounts of {grounded language,} in any natural language, will acquire the ability to comprehend and generate syntactically and semantically correct sentences in the same language --- definitions of all these terms forthcoming. 

One important remark is in order:  By designing such a system, we are not articulating a scientific theory about the precise way in which language is implemented in the human brain --- a theory to be tested by experiments on human subjects.  The artifact we create is a \emph{proof of concept,} an existence theorem stating that something akin to a language organ can be put together with basic neuroscientific materials which can be plausibly delivered by a biological developmental apparatus.  We believe that this has not been done before.  But, having said that, we have taken care that aspects of the system we present here are consistent with the consensus in beurolinguistics about the nature of the language organ, \emph{wherever such consensus exists;}  we point out instances of such convergence throughout the paper.

\def\name{{\sc nemo}}
\section{The Model}
We next turn to discussing the \emph{neural model}, henceforth referred to as \name, that we use to build our neuromorphic language organ.   Neuron biology \citep{Kandel} is rich and complex --- there are apparently thousands of different types of neural cells, hundreds of kinds of neurotransmitters, and complex and very partially understood mechanisms by which axons grow and synapses are created and synaptic ``weights'' (if one assumes that such a parameter exists) change through plasticity. It is impossible to capture everything we know in neuroscience by a model of the brain that is useful for our purposes. Our desiderata for \name\ are these:  

\begin{itemize}
    \item We want the model to be in basic agreement with what we know in neuroscience --- for example, \emph{it should not entail backpropagation.}
    \item We want it to be simple and elegant, mathematically rigorous, and amenable to mathematical proof of its properties.  
    \item Importantly, we need to simulate it efficiently, if approximately, at the scale of tens of millions of neurons and trillions of synapses. 
\end{itemize}
\name\  is very much influenced by the Assembly Calculus (AC) \citep{PNAS}, a model that was proposed a few years ago as a simplified though realistic mathematical description of brain computation, and capturing a few of the most established principals in neuroscience:  the brain is a finite collection of \emph{brain areas}, each with distinct cytological and functional properties. Individual neurons fire when they receive sufficient excitatory input from presynaptic neurons, and firing is an atomic operation.  Synapses between neurons in the same area are essentially \emph{random;} \name\ assumes the strong randomness of Erd\H os--Renyi random graphs denoted by $G_{n,p}$ \citep{Erdos60}, where all pairs of different neurons have the same probability $p$ of being connected, independently. While it is known that the randomness of synaptic connectivity is more complex than $G_{n,p}$ and influenced by locality and neuron type, see for example \citet{nonrandom}, the randomness of $G_{n,p}$ is a robust and productive assumption --- for example, alternative models of randomness based on locality bring about very similar behaviors.  Certain pairs of different brain areas can be connected by fibers of axons, in either or both directions, and this results in random \emph{bipartite} connectivity between the neurons in these areas.  

It is well known that a large minority of neurons in the mammalian brain are \emph{inhibitory} --- or \emph{GABA-ergic neurons}, as the most common type is called, or \emph{interneurons} --- and that inhibition serves two distinct functions:  \emph{Local inhibition} establishes in each area excitatory-inhibitory balance (EI balance), keeping the number of spiking neurons to a fixed fraction thus preventing seizures. Importantly, in \name\  local inhibitory neurons are not modeled explicitly; their effect is captured by the \emph{$k$-cap operation} explained below.

It is assumed in \name\  that \emph{all neurons spike in synchrony}, in distinct time steps --- implicitly assumed to run at approximately 50 Hz in the brain.  This is a necessary assumption for making \name\  mathematically tractable (so its properties can be proved analytically) and susceptible to efficient simulation. This synchrony assumption is definitely unrealistic: It is well known in Neuroscience that neuron spiking is asynchronous.  However, this assumption is \emph{not distortive:} It has been established through simulations of asynchronous neural models that the basic behaviors of the AC and \name\  are maintained in those models, see for example \citet{IsonMaassPap}.  

At each step, which neurons spike? It is asumed that, in each area, $k$ of the area's $n$ neurons fire, where $k$ is a number much smaller than $n$ --- think of it as the square root of $n$. In particular, the $k$ neurons that received the largest synaptic input from presynaptic neurons --- in the same area or in other areas --- are selected to spike.  This is the $k$-cap operation (or $k$-winners-take-all), the mechanism through which the excitatory-inhibitory balance of each area, effected by its local inhibitory neurons, is captured.  It is a productive simplification of the underlying process, in which the initial firing of many excitatory neurons excites the local inhibitory population (reacting much faster than their excitatory counterparts), which fire, inhibit many of the excitatory neurons, in return fewer inhibitory neurons fire, in an oscillation that quickly converges to the excitatory-inhibitory balance modeled by the $k$-cap.

Finally, \name\ features a simple version of \emph{plasticity}. Plasticity, the ability of neural systems to incorporate the organism's experiences, mostly through changes in synaptic weights, is a fundamental characteristic of brains, considered the basis of all learning.  There are many kinds of plasticity, and new kinds are discovered all the time; here we assume the most basic kind of Hebbian plasticity: If neuron $i$ spikes at time $t$, neuron $j$ at time $t+1$, and there is a synapse from $i$ to $j$, then the {\ weight} of this synapse, originally one, is multiplied by $1+\beta$, where $\beta>0$ is a plasticity parameter, typically $5-10\%$. There are more complex and biologically accurate models of plasticity (such as STDP); however, simulations show that the simple Hebbian version adopted in \name\ is not inaccurate in any essential way \citep{constantinides2021effects}.

We now have all ingredients of \name\  required to describe the \emph{dynamical system} that carries out brain computation. We start with a finite number of brain areas named $A,B,\ldots$, any pair of which may or may not be connected to one another through a fiber. One area $I$ is called the \emph{input} area; representations of stimuli in this area typically initiate the computation.  Each area has $n$ excitatory neurons, and at each step precisely $k$ of these through the $k$-cap operation. The neurons of each area are interconnected by a $G_{n,p}$ directed graph of synapses, where $p$ is a second parameter of the model (typically between $0.001$ and $0.01$). To summarize, the equations of the dynamical system are as follows:
\begin{itemize}
\item (State) the state of the system at time $t$ consists of, for each neuron $i$, a bit $f_i^t \in \{0,1\}$ denoting whether or not $i$ fires at time $t$, and the synaptic weights $w_{i,j}^t$ for all synapses $(i,j)$. 
\item (Synaptic input) $I_i^t$, the synaptic input of neuron $i$ at time $t$, is computed as $I_i^t = \sum_{(j,i)\in E~:~f_j^t=1} w_{j,i}$;
\item ($k$-cap) for $i$ in area $A$, $f_i^{t+1}=1$ if $I_i^t$ is in the top-$k$ of $\{I_j^t~:~j \in A\}$; 
\item (Plasticity) for each synapse $(i,j)$, $w_{i,j}^{t+1} = w_{i,j}^t(1+f_i^t f_j^{t+1} \beta)$.
\end{itemize}

Although not used explicitly in the main algorithm of this paper, our \name\ has another type of \emph{long-range interneurons}, or \emph{LRIs}, a feature absent in the AC: LRIs are distinct populations of inhibitory neurons, extrinsic to the brain areas, which have \emph{inhibitory} synaptic connections to certain brain areas or other LRIs (all other synapses in \name\ are excitatory), and have excitatory connections \emph{from} certain brain areas.  LRIs can be thought of as the \emph{control elements} of brain computation, and are crucial in making \name\ a hardware language capable of universal computation. LRIs are well attested in the neuroscience literature \citep{JinnoLRI,MelzerLRI}; in particular, there is evidence that they are necessary for establishing the $\gamma$ rhythm of the brain thought to be coterminous with brain computation \citep{RouxBuzsaki}.  LRIs rectify a marked weakness of the AC: Computation in the Assembly Calculus is represented in \citet{PNAS} by Python-like programs with variables, conditionals and loops.  It is unclear how these AC programs have evolved or how they are deployed in development, where they are stored, or how they are loaded and interpreted in the brain. LRIs replace these programs by a simple and biologically plausible framework. We discuss their use in extensions and future directions of our model, particularly for syntax, in Section \ref{open}.

\subsection*{The power of the model} 
At first glance, \name\ as described above appears to be extremely simple; however, powerful behaviors can be accomplished in this framework. One important example, studied in \cite{PNAS}, is called \emph{projection}. Let $A$ and $B$ be two areas (with a fiber from $A$ to $B$) and suppose there is a fixed set $a \subset A$ of $k$ neurons in $A$ that fires into $B$ at each time step. This setup is very simple and important: it models a \emph{fixed stimulus} firing into a brain area. 

How does the system evolve? At $t=1$, $a$ fires, resulting in some $k$-cap set of neurons $b_1$ in $B$. At $t = 2$, $a$ \emph{and} $b_1$ both fire into $B$, resulting in a some other $k$-cap $b_2$ in $B$, and so on for $t=3,4,\ldots$. A priori, it is not clear that the $b_1,b_2,b_3,\ldots$ converge, because as new neurons in $B$ fire, they might recruit more new neurons in $B$. Without plasticity (i.e., $\beta = 0$), the $b_t$ do not converge. However, as confirmed in both experiment and proof in \cite{PNAS}, for $\beta > 0$ the $b_t$ do converge to a stable set $b \subset B$. In particular, after some time step $\tilde{t}$, firing $a$ will reliably activate $b$ (by activate we mean that $b$ fires as the next time step), just as firing any reasonably-sized subset of $b$ inside $B$ also activates all of $b$. Such a stable set of neurons is called an \emph{assembly} or \emph{ensemble}, and the assembly $b$ is called the \emph{projection} of $a$ into $B$. 

There is a substantial consensus that highly interconnected sets of neurons that fire together (called assemblies, ensembles, engrams) are the fundamental unit underlying cognitive mechanisms \cite{Buzsaki:10}, and the Assembly Calculus was proposed as a model that explains and models the emergence and dynamics of assemblies. \name\ has several other operations: (1) merge, which is the formation of an assembly in an area when \emph{multiple} areas fire into it, (2) reciprocal project, when two assemblies are connected into each other both ways, and (3) sequence formation, that is a chain of projections from $A$ to $B$ that memorizes the order of projection. While the model is certainly an abstraction of neuronal activity, it is based on sound neurobiological principles, and each of these operations is a plausible abstraction of complex neural processes that are thought to underlie cognition.

\subsection*{Language in the Brain} \label{neurolinguistics}
When it comes to language in the brain, much less is known with certainty than about neuron cellular dynamics; see \citep{kemmerer,friederici,brennan2022language} for recent books on the subject. Still, it is impossible to survey the entire field. Here, we summarize the state of our knowledge of the language organ most pertinent to this work.

The language organ carries out two main functions: speech production and speech comprehension. There is a broad consensus that, in the systems responsible for both functions, there exist abstract representations for each word in the language within a centralized lexical area; this area can be thought of as a hub-like interface between the phonological subsystem and the semantic representations of each word. Though not uncontroversial, there is also growing evidence that these representations are \emph{shared} between production and comprehension systems --- they are believed to reside in the mid and mid-posterior MTG \cite{Indefrey2004}. On the other hand, the \emph{semantics} of nouns and verbs are represented in a distributed way across many brain areas, many at the periphery of the motor, visual, and other sensory cortex, and the aforementioned {word representations} are richly connected to these areas \cite{Martin2007, Kiefer2012, GallantSemantics}. 

Nouns and verbs differ in some of the context areas with which they are most strongly connected. Parts of the motor cortex are much more strongly involved in the processing of verbs than that of nouns (namely the PLTC and the pSTS subarea), whereas a different part of the motor context is more active in the processing of nouns involving action (i.e., tools and limbs) than verbs; see \citet{Gennari2012} and \citet{Watson2013} for surveys. Furthermore, areas of the motor cortex that are activated in response to perceiving someone \emph{else} perform an action, known as \emph{mirror cells,} are activated much more for verbs than for nouns; for a review on motor and mirror area recruitment for verbs vis-a-vis nouns, see \citet{KemmererGonzalez2010, Fernandino2010}. In addition to involving different context areas, it is also known that there is a separation between the systems for the perception and generation of nouns and that of verbs, and there is growing evidence that this may, at least partly, be because noun and verb lexical representations reside in different subparts of the mid and mid-posterior MTG, that is, the lexical area \cite{Matzig2009, Vigliocco2011, Kemmerer2012}. Our language organ model will reflect these principles by having separate lexical areas for nouns and verbs, and featuring contextual that are connected {\em exclusively} to each of the noun and verb areas --- in addition to many shared context areas.

Neurolinguists also strongly suspect that there is an abstract phonological representation for each word (in an area called Spt) that is connected to the word's centralized lexical representation, and that the same representation is used in both perception and production \cite{Hicock2003, OKADA2006112}. This representation can be thought of as containing implicit representations of sequences of phonemes and interfacing to the sensorimotor subsystems for the perception and production of these words. In our work, we abstract away phonological processing and acquisition, and will have a special phonological input/output area that is shared by production and perception.

To summarize, we have the following simplified picture of language in the brain: each word has a root representation in a lexical hub area, likely within different sub-areas for nouns and verbs, which is connected to a phonological representation of the word --- representations which are used both for recognizing and for articulating words.  The lexical hubs are richly connected to many sensory and semantic areas across the brain through which the many complex shades of meaning and nuances of a word are represented; crucially, nouns and verbs have strong connections to different context areas. 

% Lastly, the following facts are less important for our main goal of modeling the acquisition of word semantics for nouns and verbs, but are important nonetheless and come up in the our model extensions: (a) in multilingual speakers, noun and verb hubs/lemmas reside in roughly the same areas; (b) functional categories --- words such as \emph{the} and {I} --- though produced much later, begin to be perceived as early as other categories, and their representations are highly localized.

\subsection*{Psycholinguistic theories}
The most important comparisons to our work are the existing psycho- and neurolinguistic models of language processing. Among the most influential and established theories are the Lemma Model for production \cite{LeveltLemma}, the Dual Stream Model of language perception \cite{HickockPoeppel}, and the Hub-and-Spoke model for the semantic representation of words \cite{RalphHubSpoke}. While there is much debate regarding in which ways these models can be combined, they all have in common the basic consensuses, or near-consensuses, outlined in the previous section. One important contribution of our work is that it constitutes a \emph{concrete, neuronal implementation} of the underlying common core of these three mainstream models of language processing. That is, whereas at the highest level the Lemma Model posits the existence of word lemmas connected to phonological representations and a hub-and-spoke like semantic network, and the Dual Stream Model predicts a lexical interface between the integration of phonological input and the semantic and syntactic features of a word, our work \emph{fully implements}, in terms of realistic stylized neurons, the basic underlying mechanisms of these models. Importantly, our model explains how the \emph{lexical representations} common to these models can be acquired from grounded input. 

\paragraph{A toy language.} We will shortly define a language organ in \name\ that will learn from sentences of a toy language with $l$ nouns and $l$ intransitive verbs, where $l$ is a small parameter that we vary in our experiments. We denote the combined lexicon as $L$. In this language, all sentences are of length two: ``cats jump'' and ``dogs eat.'' Importantly --- and this is the hard part of our experiment --- the language can have either SV (subject-verb) principal word order (as in English, Chinese and Swahili) or VS (as in Irish, Classical Arabic, and Tagalog), and our model should succeed in either scenario.

\section{The Language Organ} 
\begin{figure}
    \centering
    \includegraphics{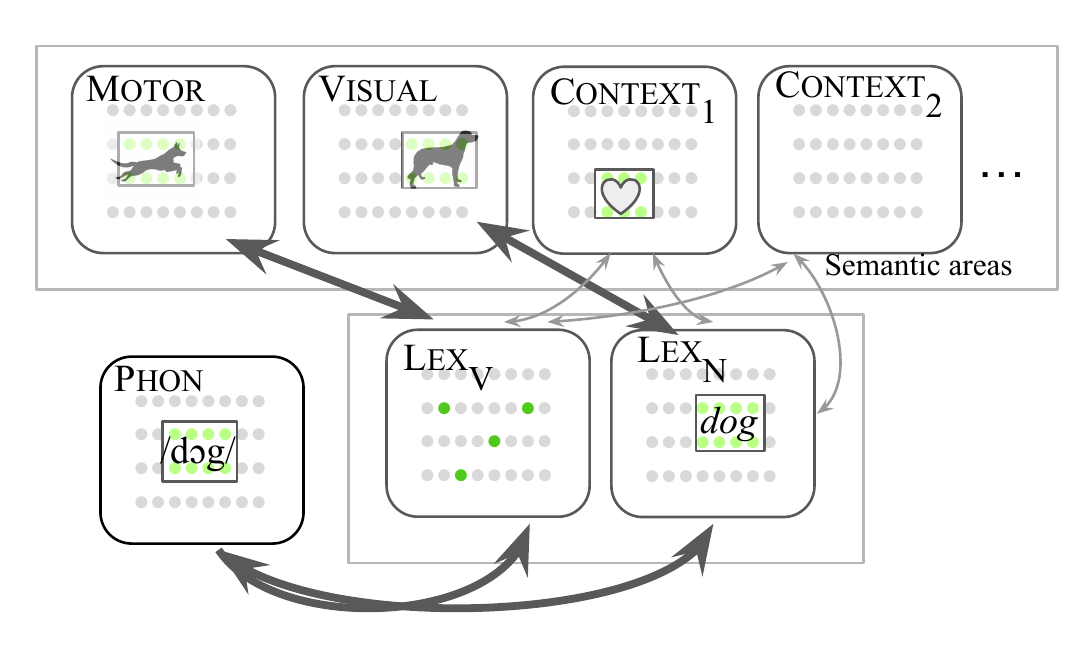}
    \hfill
    \caption{The architecture of the language organ $\mathcal{O}$ in the \name\ model of neuronal computation. This example show the state of a trained $\mathcal{O}$ after hearing the word \emph{dog} in a grounded setting when the listener also sees a dog jumping (this could be part of a sentence like ``the dog jumps"). The corresponding assemblies are active in $\Visual$ (the image of a dog) and $\Motor$ (the action of jumping); assemblies can also be active in $\Context_i$ areas, representing additional semantic contextual stimuli such as an emotional affect.}
    \label{fig:organ}
\end{figure}

Our language organ, denoted $\mathcal{O}$, consists of two separate \emph{lexical areas} for nouns and verbs, $\LexN$ and $\LexV$, and an area $\Phon$ containing the phonological representations of words.  It also has several \emph{context} areas: $\Visual$ and $\Motor$ are the two basic ones, but there are several others which we denote $\Context_i$ for $i \in [C]$. ($C$, the number of additional context areas, is a parameter of the model; here $C=10$). $\Phon$ is connected through fibers with $\LexN$ and $\LexV$, whereas $\Visual$ is connected with $\LexN$, and $\Motor$ with $\LexV$. All other context areas $\Context_i$ are connected to both $\LexN$ and $\LexV$; all these connections are two-way (see Figure \ref{fig:organ}). For each word $W$, we additionally pre-select a random subset of $[C]$, representing which extra context areas are implicated for the word $W$ (for instance an olfactory area for $W=$~\emph{flower} an emotional affect area for \emph{hug}, and so on). In our experiment, this set has only one element, denoted $i[W]$.

Hearing each word $W$ by the learner is modeled as the activation of a unique corresponding assembly $\Phon[W]$ for that word in $\Phon$ for the duration of the perception of a word, that is, for $\tau$ time steps, where $\tau$ is another parameter of the model.  We further assume that our input is \emph{grounded:}: whenever a noun $W \in L$ is heard it is also seen --- that is, an assembly corresponding to the static visual \emph{perception} of the object (cat, dog, mom, etc) is active in $\Visual$, denoted $\Visual[W]$. Similarly, an assembly corresponding to the  intransitive action (jump, run, eat, etc.) in $\Motor$, denoted $\Motor[W]$ for a verb $W \in L$. \emph{These areas represent the union of the differing somatosensory cortical areas feeding into nouns and verbs covered in Section \ref{neurolinguistics}}). We also activate an assembly $\Context_{i_W}[W]$ in the extra context area corresponding to $W$. Importantly, the assemblies in the contextual areas ($\Visual,~\Motor$ and the $\Context_i$) are activated throughout the perception of the entire sentence (that is, $\tau~\times~\big |$~sentence~$\big|$  steps), \emph{not} just when the corresponding word is perceived.  This corresponds to the fact that the learner perceives the sentence as a whole, associated with the world-state perceived that moment through shared attention with the tutor.

Effectively, the above means that in our experiment, whereas $\LexN$ and $\LexV$ are pristine \emph{tabulae rasae}, areas with random connectivity devoid of special structure, $\Phon$ is pre-initialized with assemblies for each word in the lexicon; $\Visual$ has assemblies for each noun, as does $\Motor$ for each verb. This reflects that we seek to model the acquisition of highly grounded, core lexical items, and are abstracting away phonological acquisition --- which is of course a highly interesting direction in its own right. These lexical items are acquired before more abstract nouns and verbs (such as \emph{peace} and \emph{explain}) that may require a variant of this representation scheme.   We are confident that appropriate extensions of our basic model will handle abstract language --- see Section \ref{open} for a discussion of this and other extensions.

To summarize, a sentence $s = W_1 W_2$ of our language in the SV setting (the VS setting is analogous) is input into $\mathcal{O}$ as follows: the corresponding assemblies in all the context areas, that is $\Visual[W_1]$, $\Motor[W_2]$, $\Context_{i[W_1]}$ and $\Context_{i[W_2]}$ fire for $t-1 \in [2\times \tau]$, while $\Phon[W_1]$ fires for $t-1 \in [\tau]$, and then $\Phon[W_2]$ fires for $t-\tau-1 \in [\tau]$. We will denote these steps of the dynamical system by the shorthand Feed($s$).

\section{The learning experiment} \label{experiment}
We first select the parameters $n,k,p,\beta$ (which may vary across different areas); $l$ (the lexicon size), $\tau$ (how many times each word fires), and $C$ (the number of extra context areas). To \emph{train} $\mathcal{O}$, we generate random sentences $s_1,s_2,\ldots$ in our toy language, executing Feed($s_i$) for each $s_i$.

Our experiments reveal that, for varying settings of the parameters (such as $n=10^6, k=10^3, \beta =0.1, l=5, \tau=2$), after some number of training sentences the model accomplishes something interesting and nontrivial, and \emph{necessary for language acquisition:} it forms assemblies for nouns in $\LexN$ but not in $\LexV$,  assemblies for verbs in $\LexV$ but not in $\LexN$\footnote{To see why this is highly nontrivial, the reader is reminded that this is done in the absence of knowledge of whether, in the language being learned, subject precedes verb or the other way around.}, and in addition, the assemblies in these areas are reliably connected to each word's corresponding assemblies in $\Phon,\Motor$, and $\Visual$, and also reasonably well connected to the other context areas. In other words, and in a concrete sense, the model has learned which words are nouns and verbs, and has formed correct semantic representations of each word.  %We will now define the details of this concretely, and then see how $m$ varies for different settings of the parameters (for instance, lower plasticity coefficient $\beta$ necessitates more example sentences for these stable representations to form, etc.)

We say that an experiment \emph{succeeded} after $m$ training sentences if we have that for each word $W \in L$, the resulting \emph{ synaptic weights} of $\mathcal{O}$  satisfy properties $P$ and $Q$. Property $P$ captures a kind of {\em production} ability --- that is, ability to go from semantic representations to phonological form, much like the mapping from lemma to lexeme in psycholinguistics; properties $Q$ guarantee that a stable representation for each word is formed in the word's correct area --- $\LexN$ or $\LexV$ --- and not in the other area.

We start by defining the $P$ property: A noun (respectively, verb) $W$ satisfies property $P$ if firing $\Visual[W]$ (resp.~$\Motor[W]$) and $\Context[i[W]]$ activates via $\LexN$ (resp.~via $\LexV$) almost all of the representation $\Phon[W]$; in our tests, we define ``almost'' as least 75\% of the cells in that assembly. We say the experiment satisfies $P$ if every word satisfies the $P$.
%Similarly a verb $W$ satisfies $P$ if firing $\Motor[W]$ activates almost all of $\Phon[W]$. %Property $P_2$ ??? in addition to firing $\Visual[W]$ and $\Motor[W]$ we co-fire some subset of the $\Context_i[W]$. 
%We say the experiment satisfied $P$ if all nouns satisfy $P_1$ and all verbs satisfy $P_2.

For the $Q$ properties, suppose $W$ is a noun and that $\Phon[W]$ fires once. Let $\nu$ be the resulting $k$-cap in $\LexN$, and $\mu$ the resulting $k$-cap in $\LexV$. The properties $Q_i$ are defined as follows.
\begin{enumerate}
\item $Q_1$: the synaptic input into $\nu$ is \emph{greater} than that into $\mu$ by a factor of two. 
\item $Q_2$: if we fire $\nu$, it activates $\Phon[W]$ and $\Visual[W]$; whereas if we fire $\mu$, it does not activate any of the predefined assemblies in $\Phon$ or $\Motor$.
\item $Q_3$: if we fire $\nu$, it activates $\nu$ within $\LexN$ itself; whereas if we fire $\mu$, the next $k$-cap in $\LexV$ has small overlap with $\mu$ (less than 50\%).
\end{enumerate}
If $W$ is a verb, the $Q_i$ are defined as above but swapping noun with verb, and $\Motor$ with $\Visual$. Intuitively, the $Q_i$ capture that \emph{a stable hub representation of each word has been formed in the correct part-of-speech lexical area for that word}. The experiment satisfies $Q$ is every word satisfies the $Q_i$.

\textbf{Results.} \label{results}
%Figures  discussion of experimental results.
We run our \name~-based language organ with a variety of parameters with random sentences until success, that is, until $P$ and $Q$ are satisfied, and report the resulting training time. Despite representing a dynamical system of \emph{millions of neurons and synapses}, the system converges and yields stable representations (satisfying $P$ and $Q$) for reasonable settings of the parameters.

\begin{figure}
    \centering
    \begin{subfigure}[b]{0.49\textwidth}
         \centering
         \includegraphics[width=\textwidth]{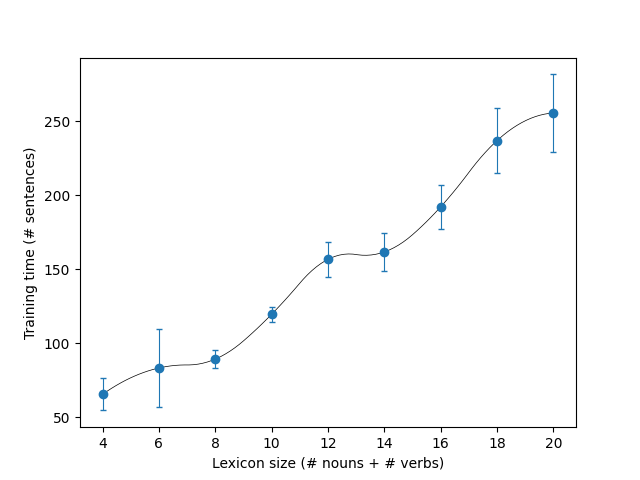}
         \caption{Increasing lexical size}
         \label{fig:lex}
    \end{subfigure}
    \hfill
    \begin{subfigure}[b]{0.49\textwidth}
         \centering
         \includegraphics[width=\textwidth]{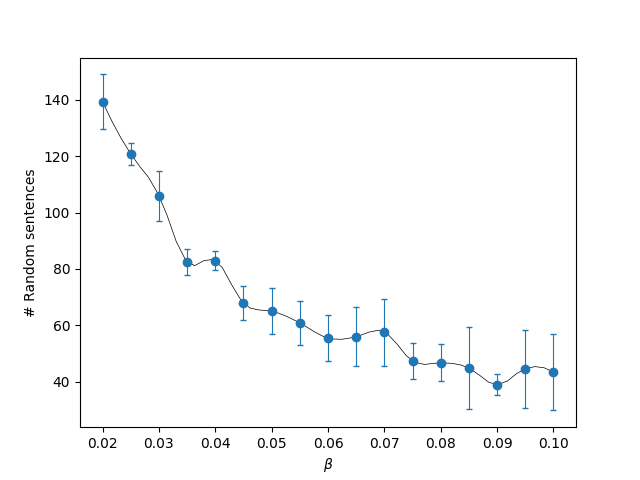}
         \caption{Varying beta}
         \label{fig:betas}
    \end{subfigure}
    \vskip\baselineskip
    \begin{subfigure}[b]{0.49\textwidth}
         \centering
         \includegraphics[width=\textwidth]{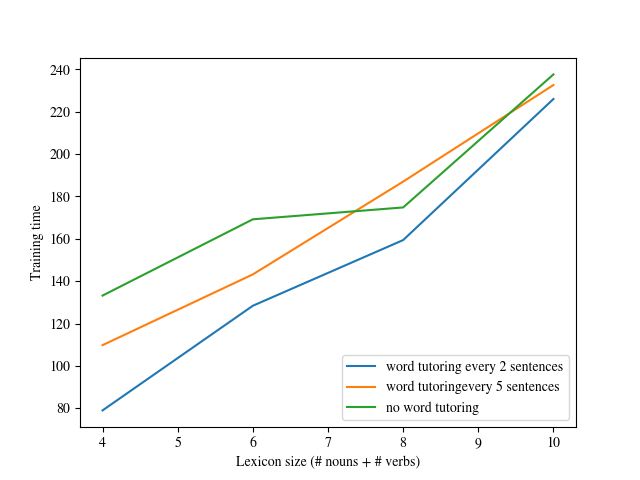}
         \caption{Mixing in individual word tutoring}
         \label{fig:tutoring}
    \end{subfigure}
    \caption{Results of our experiments. In (a) the learning experiment of section \ref{experiment} is performed for varying sizes of the lexicon, revealing a linear trend ($n=10^5,p=0.05,\beta=0.06,k_{\LexN} = k_{\LexV}=50,k_{\Context_i}=20$, other areas $k=100$, $C=20$ and $\tau=2$). In (b) the learning experiment is repeated for varying $\beta$ and $C=0$, always for a lexicon of size $4$. In (c) we run learning experiments as in (a) --- green --- along with two variants, one in which a round of individual word tutoring is performed after every $2$ random sentences (blue), and another every $5$ random sentences (green): individual word tutoring decreases training time significantly, particularly when a smaller set of words is taught at a given time. (a)-(c) were performed with both a NV and NV word orders with similar results; NV results are shown here and both are available in the supplementary data. Each experiment is repeated 5 times; means and standard deviations are reported. }
    \label{fig}
\end{figure}

The results are summarized in Figure \ref{fig}, where we see that the number of training sentences grows roughly linearly with the lexicon, or number of words acquired. While the number of training sentences may appear somewhat large, there are a few points to keep in mind. Our model describes the acquisition of one's ``first words", the most contextually rich and consistent, for which 10-20 overhead sentences per word does not seam unrealistic. Furthermore, to our knowledge ours is the first simulation of a non-trivial part of language acquisition performed entirely in a bioplausible model of neurons and synapses. Nevertheless, reducing the number of training sentences is a crucial goal of this line of research: we propose a heuristic for this in the following subsection, and discuss ideas for future research in Section \ref{open}. We also experiment running the model with varying $\beta$ (the plasticity parameter) revealing roughly inverse-exponential acceleration of the rate of convergence to stable representations with increasing $\beta$. In experiments with or without extra context areas, the training time remains roughly the same. See Figure \ref{fig} for details.

\subsection*{Individual word tutoring} 
Our model is able to learn word semantics from \emph{full} sentences, without ever being presented isolated words. While it is known that children can  acquire language in this way, in our experiments the number of sentences required is rather large, and scales linearly with the size of the lexicon. An important problem for our theory is to understand how to reduce this size, especially to model later stages of acquisition, since humans acquire language from small amounts of data.  We believe this is done two ways: At the early stages with \emph{individual word tutoring}, and at later stages through {\em functional words} (see the next section).   To test individual word tutoring, after every fixed number of sentences we randomly select a single word $W \in L$ and fire $\Phon[W]$ and its contextual areas for some $\tau$ time-steps. We find that this greatly decreases the total training \emph{time}. %(measured as number of words, which is linear to the number of firing rounds of $\mathcal{O}$; see Figure \ref{fig:tutoring}. 
In particular, at early stages of acquisition, individual word tutoring reduced the training time by over 40\%.

\section{Future Work} \label{open}
% Which of these to include for the NeurIPS submission
\paragraph{Multilinguality} 
We believe that our model can be extended to handle \emph{multilinguality} by adding an additional area $\Lang$, connected into $\LexN$ and $\LexV$. Like the contextual areas, $\Lang$ would have several assemblies, one for every language the multilingual child is exposed to, with strong input into $\LexN$ and $\LexV$. For learning to succeed in the sense of Section \ref{experiment}, separate assemblies for each concept in each language must form in the lexical areas; we expect that this will require more training time --- reflecting the fact that multilingual children may begin to speak later than monolingual children \citep{bilingualism}. 

\paragraph{Functional words and faster learning.}
Functional words are words in closed lexical classes that have limited semantic content but have important syntactic roles (such as English prepositions, determinants, etc.); more broadly, functional categories include morphemes and inflectional paradgims of this type (e.g. the possessive marker ``'s", the adverbializer ``-ly" and so on). Functional categories are somewhat of a paradox: cross-linguistically, children begin to accurately \emph{produce} them much later than lexical words (verbs and nouns), but in recent decades, an explosion in language acquisition research has come to establish that young children are extremely sensitive to them, likely forming representations of them well before they can produce them, and utilizing them in many ways: to aid understanding, for learning lexical items (a word that follows ``the'' is likely to be a noun), and for bootstrapping syntax \citep{lust}. 

An important open problem is handling functional words, and, possibly, using them to accelerate word acquisition (reducing the learning times of \ref{results}, particularly important for modeling words with less contextual consistency). As a starting point, suppose we extend our language to have a mandatory article ``a" before every noun (with no semantic content), that is, in the NV version of our language, every sentence has the form ``a \textsc{Noun} \textsc{Verb}". $\mathcal{O}$. Extending the model to acquire ``a" (perhaps as a representing in an area for functional words $\textsc{Func}$) is an important goal; then, it can be used to quickly identify any following word as a noun (i.e., forming an initial representation in $\LexN$).

\textbf{Abstract words and contextual ambiguity.}
Currently, our model of grounded context is rather simplistic: we assume only \emph{object nouns} and \emph{action verbs}, we have two areas that are specific to each kind of input, and several other unspecified contextual areas that fire randomly when we hear the word. Eventually, we would like to be able to handle abstract words like ``disagreement" and ``aspire". Extending our model, in particular its representation of semantics, to handle such words is one of our main future directions.

\paragraph{Generation and Syntax.}
\emph{Perhaps the most important direction left open by our work is syntax.} As a first step, we want the model to learn whether the toy language has NV or VN order. Concretely, this would entail the following experiment: after exposure to some number of random sentences (as in the current model), we can \emph{generate} sentences by activating the assemblies in contextual areas corresponding to every word in the sentence, and, letting the dynamical system run, it will fire the assemblies in $\Phon$ in the correct order of the language (NV or VN). This itself is but a small piece of syntax; transitive verbs and object would be the next step, which we believe can be carried out by modest, and hardly qualitative, extensions of our setup and methods. 

\section{Conclusion}
We have defined and implemented a dynamical system, composed of millions of simulated neurons and synapses in a realistic but tractable mathematical model of the brain, and in line with what is known about language in the brain at a high level, that is capable of learning representations of words from grounded language input. We believe this is a first and crucial step in neurally plausible modeling of the language organ and of language acquisition. We have outlined a number of future directions of research, within the reach of our approach, that are necessary for a complete theory of language in the brain. 

%\section*{References}
\bibliography{neurips2023}
\bibliographystyle{acl_natbib}

%%%%%%%%%%%%%%%%%%%%%%%%%%%%%%%%%%%%%%%%%%%%%%%%%%%%%%%%%%%%

\end{document}